\documentclass{article}

\PassOptionsToPackage{numbers, compress}{natbib}
 \usepackage[dblblindworkshop, final]{neurips_2025}
\workshoptitle{CauScien: Uncovering Causality in Science}

\usepackage[utf8]{inputenc} 
\usepackage[T1]{fontenc}    
\usepackage{hyperref}       
\usepackage{url}            
\usepackage{booktabs}       
\usepackage{amsfonts}       
\usepackage{nicefrac}       
\usepackage{microtype}      
\usepackage{xcolor}         
\usepackage{amssymb}
\usepackage{amsmath}
\usepackage{amsthm}
\usepackage{algorithm}
\usepackage{algpseudocode}
\usepackage[pdftex]{graphicx}
\usepackage{tikz}
\usepackage{ulem}
\newtheorem{theorem}{Theorem}
\newtheorem{lemma}{Lemma}

\newtheorem*{remark}{Remark}
\newtheorem{definition}{Definition}
\newtheorem{assumption}{Assumption}

\usepackage{listings}
\usepackage{cancel}

\usepackage{xcolor}
\definecolor{codegreen}{rgb}{0,0.6,0}
\definecolor{codegray}{rgb}{0.5,0.5,0.5}
\definecolor{codepurple}{rgb}{0.58,0,0.82}

\lstdefinestyle{mystyle}{
    commentstyle=\color{codegreen},
    keywordstyle=\color{magenta},
    numberstyle=\tiny\color{codegray},
    stringstyle=\color{codepurple},
    basicstyle=\ttfamily\footnotesize,
    breakatwhitespace=false,         
    breaklines=true,                 
    captionpos=b,                    
    keepspaces=true,                 
    numbers=left,                    
    numbersep=5pt,                  
    showspaces=false,                
    showstringspaces=false,
    showtabs=false,                  
    tabsize=2
}

\usetikzlibrary{shapes,decorations,arrows,calc,arrows.meta,fit,positioning}
\tikzset{
    -Latex,auto,node distance =1 cm and 1 cm,semithick,
    state/.style ={ellipse, draw, minimum width = 0.7 cm},
    point/.style = {circle, draw, inner sep=0.04cm,fill,node contents={}},
    bidirected/.style={Latex-Latex,dashed},
    el/.style = {inner sep=2pt, align=left, sloped}
}

\title{One-Shot Multi-Label Causal Discovery in High-Dimensional Event Sequences}
\author{
Hugo Math$^{1,2}$ \quad Robin Sch{\"o}n$^{2}$ \quad Rainer Lienhart$^{2}$ \\
$^1$ BMW Group \\
$^2$ Chair for Machine Learning and Computer Vision, Augsburg University \\
Augsburg, Germany \\
\texttt{hugo.math@bmw.de} \quad \texttt{robin.schoen@uni-a.de} \quad \texttt{rainer.lienhart@uni-augsburg.de}
}

\begin{document}
\maketitle

\begin{abstract}
Understanding causality in event sequences with thousands of sparse event types is critical in domains such as healthcare, cybersecurity, or vehicle diagnostics, yet current methods fail to scale. We present OSCAR, a one-shot causal autoregressive method that infers per-sequence Markov Boundaries using two pretrained Transformers as density estimators. This enables efficient, parallel causal discovery without costly global CI testing. On a real-world automotive dataset with \(29,100\) events and \(474\) labels, OSCAR recovers interpretable causal structures in minutes, while classical methods fail to scale— enabling practical scientific diagnostics at production scale.
\end{abstract}

\section{Introduction}
Causal discovery in event sequences is a central problem across various domains, including cybersecurity \cite{MANOCCHIO2024122564}, healthcare \cite{MedBERT, bihealth}, flight operations \cite{flight_service_cd}, and vehicle defects \cite{pdm_dtc_feature_extraction}.
These sequences of discrete events \(x_i\) recorded asynchronously over time often lead to outcomes (e.g, a diagnosed defect, a disease) denoted as labels \(\boldsymbol{y}\). While becoming more available at scale, they remain challenging to interpret beyond associations. 
Understanding \textit{why} specific events lead to particular outcomes is vital for effective diagnosis, prediction, and overall decision making \cite{liu2025learning, shtp}.

However, the majority of existing causal discovery methods remain computationally intractable in high dimensions \cite{cd_temporaldata_review, hasan2023a} with thousands of different nodes. Additionally, practitioners frequently reason about causality \textit{within individual unknown sequences}. For instance, ''what series of events captured by diagnostics led to this vehicle failure''. 

We aim to solve this in a one-shot manner: given only a single unknown sequence of observed events, we directly infer the causal structure explaining its outcomes, without needing multiple repetitions or large aggregated datasets. Specifically, we seek to extract, for each label, the minimal set of causal events—its Markov Boundary.

In this work, we introduce OSCAR: the first \uline{O}ne-\uline{S}hot multi-label \uline{C}ausal \uline{A}uto\uline{R}egressive discovery method. It leverages two Transformers \cite{tf} as density estimators to extract a compact interpretable subgraph with quantified causal indicators between events and labels, providing better explainability. 
Unlike traditional causal discovery methods that suffer label cardinality-dependent time complexity \cite{feature_selection_review, causality_based_feature_selection_2019, hasan2023a, cd_temporaldata_review}, OSCAR supports causal discovery across thousands of nodes. Thanks to its fully parallelised structure, it provides sequence-specific explainability in a matter of minutes. 
We validate our approach on a real-world vehicular dataset comprising 29,100 event types as diagnosis trouble codes and 474 labels as error patterns (EPs) representing vehicle defects \cite{math2024harnessingeventsensorydata}.

\section{Related Work}
Event sequences, such as diagnostic trouble codes in vehicles \cite{pdm_dtc_feature_extraction, math2024harnessingeventsensorydata} or electronic health records \cite{MedBERT, pmlr-v219-labach23a}, are often represented as a series of time-stamped discrete events \(S = \{(t_1, x_1), \ldots, (t_L, x_L)\}\) where \(0 \leq t_1 < \ldots \leq t_L\) the time of occurrence of event type \(x_i \in \mathbb{X}\) drawn from a finite vocabulary \(\mathbb{X}\). In multi-label settings, a binary label vector \(\boldsymbol{y} \in \{0, 1\}^{|\mathbb{Y}|}\) is attached to \(S\) and denotes the presence of multiple outcome labels drawn from \(\mathbb{Y}\) occuring at \emph{final time step \(t_L\)}. Forming a multi-labeled sequence \(S_l = (S, (\boldsymbol{y}_L, t_L))\). 

Transformers have recently shown strong performance in high-dimensional event spaces for next-event or label prediction \cite{tf, gpt, touvron2023llamaopenefficientfoundation}, including dual-architecture setups predicting both events and outcomes \cite{math2024harnessingeventsensorydata}—a structure we repurpose for causal discovery.

Neural autoregressive density estimators (NADEs) \cite{NIPS1999_e6384711} factorise sequence likelihood via the chain rule, and modern Transformer-based NADEs have been applied to causal inference \cite{density_estimator_2O21_journal_ci, im2024usingdeepautoregressivemodels} by simulating interventions or approximating Bayesian networks \cite{density_estimator_2O21_journal_ci, im2024usingdeepautoregressivemodels}.

Transformers as causal learners have been explored in sequential settings, with attention patterns interpreted as latent causal graphs \cite{tf_learns_gradient_structure, make1010019}. We extend the one-shot sequence-to-graph idea \cite{tf_causalinterpretation_neurips_2023} to multi-label causal discovery to scale to tens of thousands of event types.

Multi-label Causal Discovery seeks to identify the Markov Boundary (\textbf{MB}) of each label—its minimal set of parents, children, and spouses—such that the label is conditionally independent of all other variables given its \textbf{MB} \cite{optimal_feature_set_cd}. 
While classical constraint-based algorithms have shown success on low-dimensional tabular data \cite{constrainct_based_cd, causality_based_feature_selection_2019}, their application to event sequences with multi-label outputs remains challenging due to dimensionality, sparsity, temporal dependencies, and distributional assumptions \cite{cd_temporaldata_review, bn_np_hard, causality_based_feature_selection_2019}.

A comprehensive list of the notations, definitions, proofs, and assumptions used throughout the paper can be found respectively in Appendix \ref{sec:notations}, \ref{sec:definition}, \ref{appendix:proofs} and \ref{sec:assumptions}.  

Working with causal structure learning from observed data requires several assumptions, notably the causal Markov assumptions \cite{pearl_1998_bn} states that a variable is conditionally independent of its non-descendants given its parents. We assume the following: an event is allowed to influence any future events only (temporal precedence A\ref{assumption:temporal_precedence}), event lagged effects are contained in a fixed windows (A\ref{assumption:lagged_effects}), the transformers model perfectly the joint probability distribution of event and labels (A\ref{assumption:oracle}) and causal sufficiency (A\ref{assumption:causal_sufficiency}). A discussion on the impact of assumptions is provided in Appendix \ref{sec:6_discussion}.

\section{Methodology}\label{sec:oneshotmarkov}
\textbf{Conditional Mutual Information Estimation via Autoregressive Models.} We model each multi-labeled event sequence \(S_l = (S, \boldsymbol{y}_L)\) as a sequential Bayesian Network (Def. \ref{def:bn}), over events \(X_i \in \mathbb{X}\) and labels \(Y_j \in \mathbb{Y}\) (Fig.~\ref{fig:markov_boundary_identification}).
Our goal is to recover the \textbf{MB} of each \(Y_j\). 
Specifically, we would like to assess how much additional information event \(X_i\) occurring at step \(i\) provides about label \(Y_{j}\) when we already know the past sequence of events \(\boldsymbol{Z} = S_{<i}\). We essentially try to answer if:
\[P(Y_{j}|X_i, \boldsymbol{Z}) = P(Y_{j}|\boldsymbol{Z}) \Leftrightarrow 
D_{KL}(P(Y_{j}|X_i, \boldsymbol{Z})\|P(Y_{j}|\boldsymbol{Z})) = 0\] 
where \(D_{KL}\) denotes the \textit{Kullback-Leibler divergence} \cite{cover1999elements}. 
The distributional difference between the conditionals \(P(Y_j|X_i, \boldsymbol{Z}), P(Y_j| \boldsymbol{Z})\) is akin to Information Gain \(I_G\) \cite{quinlan:induction} conditioned on past events:

\begin{equation}
    I_G(Y_j, x_i|z) \triangleq D_{KL}(P(Y_j|X_i=x_i, \boldsymbol{Z} = z)) || P(Y_j|\boldsymbol{Z}=z)) 
\end{equation}
Which is equals to the difference between the conditional entropies \cite{cover1999elements, quinlan:induction} denoted as \(H\): 
\begin{equation}
    I_G(Y_j, x_i|z) = H(Y_j|z) - H(Y_j|x_i, z) \label{eq:info_gain}
\end{equation}

More generally, we can use the conditional mutual information (CMI) \cite{cover1999elements} as \(I\) to assess conditional independence (Def.~\ref{def:conditional_independence}). It is simply the expected value of the information gain \(I_G(Y_j, x_i|z)\)  such as:
\begin{equation}
    I(Y_{j}, X_i|\boldsymbol{Z}) \triangleq H(Y_j|\boldsymbol{Z}) - H(Y_j|\boldsymbol{Z}, X_i)\label{eq:cmi_theorique}
    = \mathbb{E}_{p(z)}[I_{G}(Y_j, X_i|\boldsymbol{Z}=z)]
\end{equation}

\begin{figure}[!t]
    \centering
    \caption{The overview of OSCAR: \uline{O}ne-\uline{S}hot multi-label \uline{C}ausal \uline{A}utoRegressive discovery. \(d\) denotes the hidden dimension, \(L\) the sequence length, \(\text{MB}_1, \text{MB}_2\) the Markov Boundary of \(Y_1, Y_2\) respectively. All green and blue areas represent parallelised operations.}
    \includegraphics[width=1\linewidth]{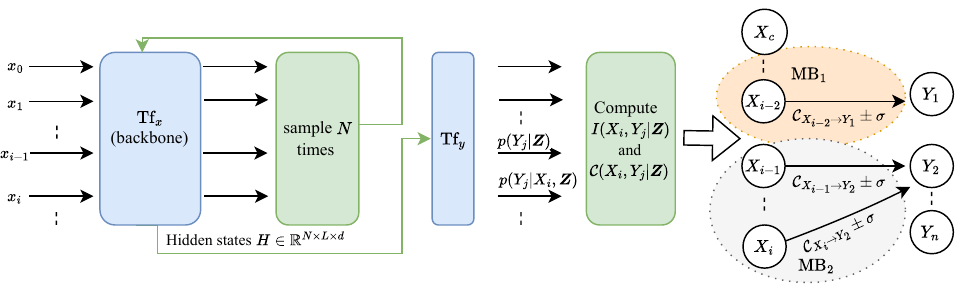}
    \label{fig:oscar}
\end{figure}

It can be interpreted as the expected value over all possible contexts \(\boldsymbol{Z}\) of the deviation from independence of \(X_i, Y_j\) in this context. To approximate  Eq.~\eqref{eq:cmi_theorique},
 a naive Monte Carlo \cite{Doucet2001} approximation is performed where we draw $N$ random variations of the conditioning set 
\(z^{(l)} = \{x^{(l)}_0, \ldots, x^{(l)}_{i-1}\}\), denoting the $l$-th sampled \emph{particle}:
\begin{align}\label{eq:cmi_approx}
\hat{I}_N(Y_{j}, X_i \mid \boldsymbol{Z}) 
&= \frac{1}{N} \sum_{l=1}^N I_G(Y_{j}, X_i \mid \boldsymbol{Z} = z^{(l)})
\end{align}
This estimator is unbiased because the contexts \(z^{(l)}\) are sampled directly from \(\text{Tf}_x\) using a proposal \(Q\) with the same support as \(P(\boldsymbol{Z})\).
Since \(I_G(Y_{j}, X_i \mid \boldsymbol{Z} = z)\) is a difference between conditional entropies (\eqref{eq:info_gain}), it is thus bounded uniformly \cite{cover1999elements} by the log of supports such as:

\[0 < I_G(Y_{j}, X_i \mid \boldsymbol{Z} = z^{(l)}) = H(Y_{j}|z^{(l)}) - H(Y_{j}|X_i, z^{(l)})) \leq H(Y_{j}) \leq \log{|\mathbb{Y}|}\]

Thus the posterior variance of \(f_i =I_G(Y_{j}, X_i \mid \boldsymbol{Z} = z^{(l)})\) satisfies \(\sigma^2_{f_i} \triangleq \mathbb{E}_{p(z)}[f^2_i(p(z)] - I^2(f_i) < +\infty\) \cite{Doucet2001} then the variance of \(\hat{I}_N(f_i)\)) is equal to \(\textit{var}(\hat{I}_N(f_i)) = \frac{\sigma^2_{f_t}}{N}\) and from the strong law of large numbers: 
\begin{align}
\hat{I}_N &\xrightarrow[N \to +\infty]{\text{a.s.}} 
\mathbb{E}_{p(z)}\!\left[ I_G(Y_{j}, X_i \mid \boldsymbol{Z}=z) \right] 
\triangleq I(f_i).
\end{align}


\textbf{Sequential Markov Boundary Recovery.}
In practice a per-label threshold \(\theta_j \approx 0\) is applied to Eq~\eqref{eq:cmi_approx} to identify conditional independence:
\begin{equation}\label{eq:cmi_epsilon}
    Y_j \not\!\perp X_i \mid \boldsymbol{Z} \quad \Leftrightarrow \quad I(Y_j, X_i \mid \boldsymbol{Z}) > \theta_j \approx 0
\end{equation}
\(\theta_j\) is dynamically computed for each label based on the mean and standard deviation of the CMI values across the sequence such that: \(\theta_j = \mu_{Y_j} + k \cdot \sigma_{Y_j}\). We analyse the effect of \(k\) in Fig.~\ref{abl:threshold_selection} as well as the effect of the proposal \(Q\) and number of particles in Appendix \ref{abl:sampling} and \ref{sec:samplingnumber}.

We reuse the two architectures introduced by \citet{math2024harnessingeventsensorydata} to perform next event prediction (\textit{CarFormer} as \(\text{Tf}_{x}\)) and next labels (\textit{EPredictor} as \(\text{Tf}_y\)) with past events \(\boldsymbol{Z} = (x_1, \cdots, x_{i-1})\)
\begin{align}
\text{Tf}_x(S_{< i}) &= \textit{Softmax}(\boldsymbol{h}^{x}_{i-1}) = P_{\theta_x}(X_i| \boldsymbol{Z}) \label{eq:prob_x} \\
\text{Tf}_y(S_{\leq i}) &=  \textit{Sigmoid}(\boldsymbol{h}^{y}_{i}) = P_{\theta_y}(Y|X_i, \boldsymbol{Z})\label{eq:prob_y}
\end{align}
Here, \(\boldsymbol{h}^{x}_{i-1}, \boldsymbol{h}^{y}_{i} \in \mathbb{R}^{d}\) are the logits produced by \( \text{Tf}_x\) and \(\text{Tf}_y\) parametrized by \(\theta_x, \theta_y\).  
\begin{theorem}[Markov Boundary Identification in Event Sequences]
\label{th:mb-recovery}
If  \(S^k_l\) a multi-labeled sequence drawn from a dataset  \(D = \{S^1_l, \cdots, S^n_l\} \subset \mathbb{S}\) where two Oracle Models \(\text{Tf}_x\) and \(\text{Tf}_y\) were trained on, then under causal sufficiency (A\ref{assumption:causal_sufficiency}), bounded lagged effects (A\ref{assumption:lagged_effects}) and temporal precedence (A\ref{assumption:temporal_precedence}), the Markov Boundary of each label \(Y_j\) in the causal graph \(\mathbb{G}\) can be identified using conditional mutual information for CI-testing. 
\end{theorem}

Intuitively, Theorem~\ref{th:mb-recovery} states that if our Transformers perfectly approximate the true joint distribution, then testing conditional mutual information at each step is sufficient to recover the Markov Boundary of each label sequentially. By induction, we prove that with bounded lagged effects of the previous events, we can restrict their causal influence and recover the correct \textbf{MB} of each label in the associated Bayesian Network (Fig.~\ref{fig:markov_boundary_identification}).

These guarantees rest on strong assumptions—causal sufficiency, oracle-quality density estimators, and bounded lagged effects. While these are unlikely to strictly hold in practice, they simplify identifiability and highlight where practical approximations may degrade performance (Appendix \ref{sec:6_discussion}). 
We particularly observed that for rare labels, the one-shot performances drastically dropped. We analyze this behavior in the experiment section in Fig.~\ref{fig:markov_len}. One must carefully evaluate \(\text{Tf}_y\) classification performance on downstream tasks using macro metrics before applying OSCAR.


\textbf{Computation.}
A key advantage of our approach is its scalability. Unlike traditional methods whose complexity depends on the event and label cardinality \(|\mathbb{X}|\) and \(|\mathbb{Y}|\) \cite{feature_selection_review}, \emph{our method is agnostic to both}. 
Fig.~\ref{fig:oscar} shows all parallelised steps on GPUs. CMI estimations are independently performed for all positions \(i \in [c, L]\), with the sampling pushed into the batch dimension and results averaged across particles using Eq.~\ref{eq:cmi_approx}. This transitions the time complexity from \(\mathcal{O}(\text{BS} \times N \times L)\) to \(\mathcal{O}(1)\) per batch. A Pytorch \cite{pytorch} implementation is given in Appendix~\ref{lst:oscar}.

\textbf{Causal Indicator.}
While deterministic DAGs reveal structural dependencies, they often obscure the \textit{magnitude} and \textit{direction} of influence between variables. Given that we can estimate conditional distributions, we define the \textit{causal indicator} \(\mathcal{C} \in [-1, 1]\) \cite{measure_of_causal_strength_oxford, Eells_1991} between an event \(X_i\) and a label \(Y_j\) under context \(\boldsymbol{Z}\) that we assume fixed for every measurement \cite{measure_of_causal_strength_oxford} as:
\begin{equation}\label{eq:causal_indic}
\mathcal{C}(Y_j, X_i; \boldsymbol{Z}) := \mathbb{E}_{p(z)}[P(Y_j =1\mid X_i=1, \boldsymbol{Z}=z) - P(Y_j =1\mid X_i=0, \boldsymbol{Z} = z)]
\end{equation}
This enables an easy interpretation, for instance, if \(\mathcal{C} < 0\), then \(X_i\) inhibits the occurrence of \(Y_j\). We employ the term causal \textit{indicator} to separate from causal strength measures, which, if using this formulation, can be problematic \cite{quantifyingcausalinfluence}. Ours serves more as an indication of the rise in label likelihood after observing a certain event, rather than a strength. The causal strength, strictly speaking, is here the CMI since it serves as a CI-test to recover the correct DAG.

\section{Empirical Evaluation}

\textbf{Comparisons.}\label{sec:settings}
Although no existing method directly targets one-shot multi-label causal discovery \cite{cd_temporaldata_review}, we benchmark OSCAR against local structure learning (LSL) algorithms that estimate global Markov Boundaries. This includes established approaches such as CMB \cite{cmb}, MB-by-MB \cite{WANG2014252}, PCD-by-PCD \cite{pcdpcd}, IAMB \cite{iamb} from the \textit{PyCausalFS} package \cite{causality_based_feature_selection_2019}, as well as the more recent, state-of-the-art MI-MCF \cite{mimcf}. 
We used a \(g4dn.12xlarge\) instance from AWS Sagemaker to run comparisons, containing 4 T4 GPUs. We used a combination of F1-Score, Precision, and Recall with different averaging \cite{reviewmultilabellearning} (Appendix ~\ref{appendix:eval}) to perform the comparisons. The code for OSCAR, \(\text{Tf}_x, \text{Tf}_y\) and the evaluation are provided anonymously \footnote{\url{https://github.com/Mathugo/OSCAR-One-Shot-Causal-AutoRegressive-discovery.git}} as well as the anonymised version of the dataset for reproducibility purposes. An Ablation of the NADEs quality is given in \ref{appendix:nades}.

\textbf{Vehicle Event Sequences Dataset.}
We evaluated our method on a real-world vehicular test set of \(n=300,000\) sequences. It contains \(|\mathbb{Y}|=474\) different error patterns and about \(|\mathbb{X}| = 29,100\) different DTCs forming sequences of \( \approx 150 \pm 90\) events. We used 105m backbones as \(\text{Tf}_x, \text{Tf}_y\) \cite{math2024harnessingeventsensorydata}. The two NADEs were not exposed to the test set during. 
The error patterns are manually defined by domain experts as boolean rules between DTCs in Eq \eqref{eq:ep_def} where \((y_1)\) is a boolean definition based on diagnosis trouble codes \((x_i)\):
\begin{equation}\label{eq:ep_def}
  y_1 = x_1 \;  \& \; (x_5 \; | \; x_8 \;) \; \& \; (x_{18}\; | \; x_{12} ) \; \& \; x_3 \; \& \; (!x_{10} \;| \;!x_{20})
\end{equation}

We set the elements of this rule as the correct Markov Boundary for each label \(y_j\) in the tested sequences. It is important to note that rules are subject to changes over time by domain experts, making it more difficult to extract the true \textbf{MB}. Moreover, there is about  \(12\%\) missing \textbf{MB} rules for certain \(Y_j\). 

\begin{table}[h]
\centering
\caption{Comparisons of \textbf{MB} retrieval with \(n=50,000\) samples, \(|\mathbb{X}| = 29,100, |\mathbb{Y}|=474\) averaged over \(6-\)folds. Classification metrics averaging is 'weighted' and shown as one-shot for OSCAR. The symbol ’-’ indicates that the algorithm didn't output the \(\textbf{MBs}\) under 3 days. Metrics are given in \(\%\).}
\begin{tabular}{lcccc}
\hline
\textbf{Algorithm} & \textbf{Precision}↑ & \textbf{Recall}↑ & \textbf{F1}↑ & \textbf{Running Time (min)}↓ \\ \hline
IAMB & - & - & - & \(>4320\) \\
CMB & - & - & - & \(>4320\) \\
MB-by-MB & - & - & - & \(>4320\) \\
PCDbyPCD & - & - & - & \(>4320\) \\
MI-MCF & - & - & - & \(>4320\) \\
OSCAR & \(\mathbf{55.26 \pm 1.42}\) & \(\mathbf{31.37 \pm 0.82}\) & \(\mathbf{40.02 \pm 1.03}\) & \(\boldsymbol{11.7}\)\\ 
\hline
\end{tabular}
\label{tab:performance_comparison}
\end{table}
Table~\ref{tab:performance_comparison} shows comparision with \(n=50,000\). We found out LSL algorithms failed to compute the Markov Boundaries within multiple days (3-day timeout), far exceeding practical limits for deployment. OSCAR, on the other hand, shows robust classification over a large amount of events \((29,100)\), especially \(55\%\) precision, in a matter of minutes.
This behaviour highlights the current infeasibility of multi-label causal discovery in high-dimensional event sequences, since it relies on expensive global CI-testings \cite{cd_temporaldata_review}. This positions OSCAR as a more feasible approach for large-scale causal per-sequence causal reasoning in production environments.

We exemplify the explainability provided by our method for the task of explaining error patterns happening to a vehicle (Fig.~\ref{fig:graph_cropped}). OSCAR’s output could directly support domain expert rule refinement in diagnostics (e.g., engineers update fault detection rules), leading to a better automation of quality processes. More examples are given in the Appendix~\ref{appendix:explaination_ex}.

\begin{figure}[!h]
    \centering
    \caption{Evolution of the One-Shot Recall, Precision and F1-Score in function of the Markov Boundary length \(|\textbf{MB}(Y_j)|\) using \(n=45969\) samples.}
    \includegraphics[width=1\linewidth]{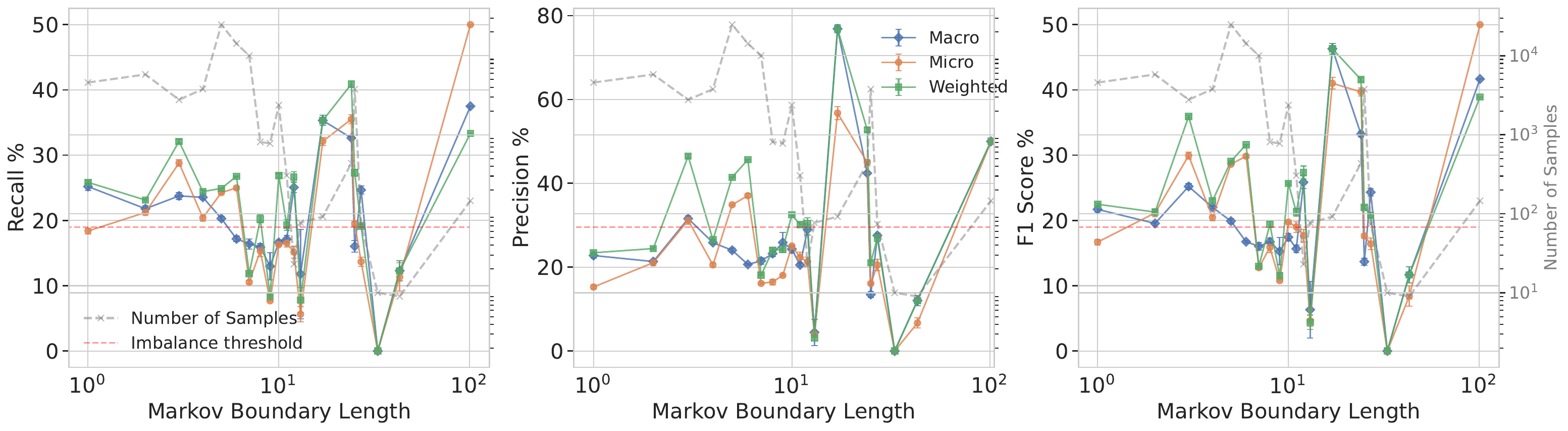}
    \label{fig:markov_len}
\end{figure}

\textbf{Markov Boundary Length.}
 Figure~\ref{fig:markov_len} reveals the classification performance depending on the number of nodes in the ground truth \(\textbf{MB}\). On the same plot is drawn in \textcolor{gray}{grey} the number of samples that each \(\textbf{MB}\) length contains (to account for imbalance). We observe that generally, a bigger \(|\textbf{MB}(Y_j)|\) does not imply a reduction in performance, highlighting the capability of OSCAR to retrieve complex Markov Boundaries in high-dimensional data. However, we observe that past a certain number of samples (imbalance threshold in \textcolor{red}{red} \(\approx 7 \times 10^2\) samples), the classification metrics are directly correlated with the number of samples per \(|\textbf{MB}(Y_j)|\). This indicates that \( \text{Tf}_y\) struggles to output proper conditional probabilities, which \emph{deteriorates the CI-test when having rare classes}. Therefore, when using OSCAR and more generally assumption A\ref{assumption:oracle}, one should carefully assess class imbalance in the pretraining phase.

\section{Conclusion}
We introduced OSCAR, a scalable one-shot causal discovery framework for high-dimensional multi-label event sequences, achieving in minutes what classical methods cannot compute in days.

We pointed out limitations, notably the oracle assumption, which might not hold for rare labels. Nevertheless, by combining pretrained autoregressive Transformers with vectorized CI-testing, OSCAR delivers, for the first time, interpretable causal graphs with quantified causal indicators for thousands of different events. This brings causal discovery closer to large-scale reasoning, making it practical for real-world sequential data.

\bibliography{math}
\appendix

\section{Notations}\label{sec:notations}
We use capital letters (e.g., \(X\)) to denote random variables, lower-case letters (e.g., \(x\)) for their realisations, and bold capital letters (e.g., \(\boldsymbol{X}\)) for sets of variables. Let \(\boldsymbol{U}\) denote the set of all (discrete) random variables. We define the event set \(\boldsymbol{X} = \{X_1, \ldots, X_n\} \subset \boldsymbol{U}\), and the label set \(\boldsymbol{Y} = \{Y_1, \ldots, Y_n\} \subset \boldsymbol{U}\). When explicitly said, event \(X^{(t_i)}_i\) represent the occurrence of \(X_i \) at step \(i\) and time \(t_i\). Similarly for \(Y^{(t_{i+1})}_{i+1}\).
\section{Definitions}\label{sec:definition}

\begin{definition}[Bayesian Network]\label{def:bn} \citet{pearl_1998_bn}
Let \(P\) denote the joint distribution over a variable set \(\boldsymbol{U}\) of a directed acyclic graph (DAG) \(\mathbb{G}\). The triplet \(<\boldsymbol{U}, \mathbb{G}, P>\) constitutes a BN if the triplet \(<\boldsymbol{U}, \mathbb{G}, P>\) satisfies the Markov condition: every random variable is independent of its non-descendant variables given its parents in \(\mathbb{G}\). Each node \(X_i \in \boldsymbol{U}\) represents a random variable. The directed edge \((X_i \rightarrow X_j)\) encodes a probabilistic dependence. The joint probability distribution can be factorized \(P(X_1, \cdots, X_n) = \prod^n_{i=1} P(X_i|X_1, \cdots, X_{i-1})\). If a variable does not depend on all of its predecessors, we can write: \(P(X_i|X_1, \cdots, X_{i-1}) = P(X_i|\text{par}(X_i))\) with 'par' the parents of node \(X_i\) such that: par\((X_i) = \{ X_1, \cdots, X_{i-1}\}\).
\end{definition}

\begin{definition}[Faithfulness]\label{def:bn_faithfulness}\citet{Spirtes2001CausationPA}. Given a BN \(<\boldsymbol{U}, \mathbb{G}, P>, \mathbb{G}\) is faithful to \(P\) if and only if every conditional independence present in \(P\) is entailed by \(\mathbb{G}\) and the Markov condition holds. \(P\) is faithful if and only if there exist a DAG \(\mathbb{G}\) such that \(\mathbb{G}\) is faithful to \(P\).
\end{definition}

\begin{definition}[Markov Boundary]\label{def:markov_boundary} \citet{optimal_feature_set_cd}.
In a faithful BN \(<\boldsymbol{U}, \mathbb{G}, P>\), for a set of variables \(\boldsymbol{Z} \subset \boldsymbol{U}\) and label \(Y \in \boldsymbol{U}\), if all other variables \(X \in \{\boldsymbol{X} - \boldsymbol{Z}\}\) are independent of \(Y\) conditioned on \(\boldsymbol{Z}\), and any proper subset of \(\boldsymbol{Z}\) do not satisfy the condition, then \(\boldsymbol{Z}\) is the Markov Boundary of \(Y\): \(\textbf{MB}(Y)\).
\end{definition}

\begin{definition}[Conditional Independence]\label{def:conditional_independence} Variables \(X\) and \(Y\) are said to be conditionally independent given a variable set \(\boldsymbol{Z}\), if \(P(X, Y|\boldsymbol{Z}) = P(X|\boldsymbol{Z})P(Y|\boldsymbol{Z})\), denoted as \(X \bot \space \space Y| \boldsymbol{Z}\). Inversely, \(X \not\perp \space \space Y| \boldsymbol{Z}\) denotes the conditional dependence.
Using the conditional mutual information \cite{cover1999elements} to measure the independence relationship, this implies that \(\text{I}(X,Y|\boldsymbol{Z}) = 0 \Leftrightarrow X \perp Y | \boldsymbol{Z}\).
\end{definition}

\section{Assumptions}\label{sec:assumptions}

\begin{assumption}[Temporal Precedence]\label{assumption:temporal_precedence}
Given a perfectly recorded sequence of events \(((x_1, t_1), \cdots, (x_L, t_L))\) with labels \((\boldsymbol{y}_L, t_L)\) and monotonically increasing time of occurrence \(0 \leq t_1 \leq \cdots \leq t_L\), an event \(x_i\) is allowed to influence any subsequent event \(x_j\) such that \(t_i \leq t_j\) and \(i < j\). Formally, the graph \(\mathbb{G} = (\boldsymbol{U}, \boldsymbol{E})\), \( (x_i, x_j) \in \boldsymbol{E} \implies t_i \leq t_j \; \text{and step} \; i < j\)
\end{assumption}
It allows us to remove ambiguity in causal directionality. By allowing for instantaneous rates (\(t_i = t_{i+1})\), our method differs from \citet{granger_causality} causal discovery. 
\begin{assumption}[Bounded Lagged Effects]\label{assumption:lagged_effects}
Once we observed events up to timestamp \(t_i\) and step \(i\) as \(\boldsymbol{Z}_{\leq t_i} = ((x_1, t_1), \cdots, (x_i, t_i))\), any future lagged copy of event \(X^{(t_i + \tau)}_i\) is independent of \(Y_j\) conditioned on \(\boldsymbol{Z}_{\leq t_i}\):
\[
Y_j \perp X^{(t_i + \tau)}_i | \boldsymbol{Z}_{\leq t_i}
\]
Where \(\tau = t_{i+1} - t_i\) is a finite bound on the allowed time delay for causal influence. 
\end{assumption}
In other words, we allow the causal influence of event \(X_i\) on \(Y_j\) until the next event \(X_{i+1}\) is observed. We note that for data with strong lagged effects (e.g., financial transactions), this might not hold well, but it is relevant for log-based and error code-based data.
\begin{assumption}[Causal Sufficiency for Labels]
\label{assumption:causal_sufficiency}
All relevant variables are observed, and there are no hidden confounders affecting the labels.
\end{assumption}

\begin{assumption}[Oracle Models]
\label{assumption:oracle} 
We assume that two autoregressive Transformer models, \(\text{Tf}_x\) and \(\text{Tf}_y\), are trained via maximum likelihood on a dataset of multi-labeled event sequences \(D_n = \{S^1_l, \cdots, S^n_l\} \subset \mathbb{S}\), and can perfectly approximate the true conditional distributions of events and labels:
\begin{equation}\label{eq:oracle}
    P(X_i|\text{Pa}(X_i)) = P_{\theta_x}(X_i|\text{Pa}(X_i)) = \text{Tf}_x(S_{< i}), \;
    P(Y_j|\text{Pa}(Y_j)) = P_{\theta_y}(Y_j|\text{Pa}(Y_j))=\text{Tf}_y(S_{\leq i})
\end{equation}

\end{assumption}
\section{Lemmas}\label{sec:lemmas}

\begin{lemma}[Identifiability of \(\mathbb{G}\)]\label{lemma:oracle_identifiability}
    Assuming the faithfulness condition holds for the true causal graph \(\mathbb{G}\). Let \(\text{Tf}_x\) and \(\text{Tf}_y\) be oracle models that model the true conditional distributions of events and labels, respectively. The joint distribution \(P_{\theta_x, \theta_y}\) can then be constructed, and any conditional independence detected from the distributions estimated by \(\text{Tf}_x\) and \(\text{Tf}_y\) corresponds to a conditional independence in \(\mathbb{G}\):
    \[
    X_i \perp_{\theta_x, \theta_y} Y_j \mid \boldsymbol{Z} \quad \implies \quad X_i \perp_{\mathbb{G}} Y_j \mid \boldsymbol{Z}.
    \]
    Where \( \perp_{\theta_x, \theta_y}\) denotes the independence entailed by the joint probability \(P_{\theta_x, \theta_y}\).
\end{lemma}

\begin{lemma}[Markov Boundary Equivalence]\label{lemma:mb_par}
In a multi-label event sequence \(S_l\) and under the temporal precedence assumption A\ref{assumption:temporal_precedence}, the Markov Boundary of each label \(Y_j\) is only its parents such that \(\forall X \in \{\boldsymbol{U} - \text{Pa}(Y_j)\},  X \perp Y_j|\text{Pa}(Y_j) \Leftrightarrow \text{MB}(Y_j) = \text{Pa}(Y_j) \). 
\end{lemma}

\section{Proofs}\label{appendix:proofs}
We provide proofs for the results described in Section \ref{sec:oneshotmarkov}

\subsection{Proof of Lemma 1}\label{proof:lemma_one_faith}
\begin{proof}
We assume that the data is generated by the associated causal graph \(\mathbb{G}\) following the sequential BN from a multi-labelled sequence \(S\). And that the faithfulness assumption holds \cite{pearl_1998_bn}, meaning that all conditional independencies in the observational data are implied by the true causal graph \(\mathbb{G}\).
Given that the Oracle models \(\text{Tf}_x\) and \(\text{Tf}_y\) are trained to perfectly approximate the true conditional distributions, for any variable \(U_i\) in the graph, we have:
\[
    P(U_i | \text{Pa}(U_i)) =
    \begin{cases}
        P(Y_j | \text{Pa}(Y_j)) = P_{\theta_y}(Y_j | \text{Pa}(Y_j)), & \text{if } U_i \in \boldsymbol{Y} \\
        P(X_i | \text{Pa}(X_i)) = P_{\theta_x}(X_i | \text{Pa}(X_i)), & \text{otherwise}.
    \end{cases}
\]
The joint distribution \(P_{\theta_x, \theta_y}\) can then be constructed using the chain rule \(P_{\theta_x, \theta_y}(X_1, \cdots, X_i, Y_1, \cdots, Y_c) = \prod^i_{k=0}P(X_k|Pa(X_k)) \prod^c_lP(Y_l|\text{Pa}(Y_l)\).
By the faithfulness assumption \cite{pearl_1998_bn}, if the conditional independencies hold in the data, they must also hold in the causal graph \(\mathbb{G}\): 
\[X_i \perp Y_j | \boldsymbol{Z} \implies X_i \perp_{\mathbb{G}} Y_j | \boldsymbol{Z}\]
Since we can approximate the true conditional distributions, it follows that:
\[X_i \perp_{\theta_x, \theta_y} Y_j |\boldsymbol{Z} \implies X_i \perp Y_j | \boldsymbol{Z} \implies X_i \perp_{\mathbb{G}} Y_j | \boldsymbol{Z}\]
Where \( \perp_{\theta_x, \theta_y}\) denotes the independence entailed by the joint probability \(P_{\theta_x, \theta_y}\).
Thus, the graph \(\mathbb{G}\) can be identified from the observational data.
\end{proof}

\subsection{Proof of Lemma 2}\label{proof:lemma_two_mb}
\begin{proof}
    Let \(<\boldsymbol{U}, \mathbb{G}, P>\) be the sequential BN composed of the events from the multi-labeled sequence \(S_l = (\{(t_1, x_1, \cdots, (t_L, x_L)\}^L_{i=1}, (\boldsymbol{y}_L, t_L))\). Following the 
temporal precedence assumption A\ref{assumption:temporal_precedence}, the labels \(\boldsymbol{y_L}\) can only be caused by past events \((x_1, \cdots, x_L)\); moreover, by definition, labels do not cause any other labels. Thus, \(Y_j\) has no descendants, so no children and spouses. Therefore, together with the Markov Assumption we know that \(\forall X \in \{\boldsymbol{U} - Pa(Y_j)\}: Y_j \perp X|Pa(Y_j)\). Which is the definition of the MB (Def. \ref{def:markov_boundary}). Thus, \(\textbf{MB}(Y_j) = Pa(Y_j)\).

\end{proof}

\subsection{Proof of Theorem 1.}
\begin{proof}\label{proof:th1}
    By recurrence over the sequence length \(L\) of the multi-label sequence \(S^k_l\), we want to show that under temporal precedence A\ref{assumption:temporal_precedence}, bounded lagged effects A\ref{assumption:lagged_effects}, causal sufficiency A\ref{assumption:causal_sufficiency}, Oracle Models A\ref{assumption:oracle} the Markov Boundary of label \(Y_j\) can be identified in the causal graph \(\mathbb{G}\). 
    
Let's define \(\mathcal{M}^L_j\) as the estimated Markov Boundary of \(Y_j\) after observing \(L\) events.

\textbf{Base Case:} \(\mathbf{L=1}\):
Consider the BN for step \(L=1\) following the Markov assumption \cite{pearl_1998_bn} with two nodes \(X_1, Y_j\). Using \(\text{Tf}_x, \text{Tf}_y\) as Oracle Models A\ref{assumption:oracle}, we can express the conditional probabilities for any node \(U\):

\begin{equation}
    P(U | \text{Pa}(U)) =
    \begin{cases}
            P(X_1) = P_{\theta_x}(X_1 | [CLS]) \; \text{if}\; U \in \boldsymbol{X} \\
            P(Y_j|X_1) = P_{\theta_y}(Y_j | X_1)\; \text{otherwise}
    \end{cases}
\end{equation}

Assuming that P is faithful (A\ref{def:bn_faithfulness}) to \(\mathbb{G}\), no hidden confounders bias the estimate (A\ref{assumption:causal_sufficiency}) and temporal precedence (A\ref{assumption:temporal_precedence}), we can estimate the CMI \ref{eq:cmi_theorique} such that iif \(I(X_1, Y_j)|\emptyset) > 0 \Leftrightarrow Y_j \not\perp_{\theta_x, \theta_y} X_1 \implies Y_j \not\perp_{\mathbb{G}} X_1\) (Lemma \ref{lemma:oracle_identifiability}).

Since we assume temporal precedence A\ref{assumption:temporal_precedence}, we can orient the edge such that \(X_1\) must be a parent of \(Y_j\) in \(\mathbb{G}\). Using Lemma \ref{lemma:mb_par}, we know that \(Par(Y_j) = \textbf{MB}(Y_j) \implies X_{1} \in \textbf{MB}(Y_j)\), thus we must include \(X_1\) in \(M^1_j \), otherwise not. 

\textbf{Heredity:}
For \(L = i\), we obtained \(M^{i}_j\) with the sequential BN up to step \(L=i\). 
Now for \(L = i+1\), the sequential BN has \(i+2\) nodes denoted as \(\boldsymbol{U'} = (X_1, \cdots, X_i, X_{i+1}, Y_j)\). Using the Oracle Models A\ref{assumption:oracle} and following the Markov assumption \citep{pearl_1998_bn}, we can estimates the following conditional probabilities for any nodes \(U \in \boldsymbol{U'}\):

\begin{equation}
    P(U | \text{Pa}(U)) =
    \begin{cases}
        P(Y_j | \text{Pa}(Y_j)) \approx P_{\theta_y}(Y_j | \text{Pa}(Y_j)), & \text{if } U \in \boldsymbol{Y} \\
        P(X| \text{Pa}(X)) \approx P_{\theta_x}(X|\text{Pa}(X)), & \text{otherwise}.
    \end{cases}
\end{equation}
By bounded lagged effects (A\ref{assumption:lagged_effects}) we know that the causal influence of past \(X_{\leq i}\) on \(Y_j\) has expired. Moreover, no hidden confounders (A\ref{assumption:causal_sufficiency}) bias the independence testing.
Finally,
using Eq.~\eqref{eq:cmi_theorique}, we can estimate the CMI such that iif
\(I( Y_j, X_{i+1}| \boldsymbol{Z}) > 0 \Leftrightarrow Y_j \not\perp_{\theta_x, \theta_y}  X_{i+1} | \boldsymbol{Z} \implies Y_j \not\perp_{\mathbb{G}} X_{i+1} | \boldsymbol{Z}\) (Lemma \ref{lemma:oracle_identifiability}).

Since we assume temporal precedence A\ref{assumption:temporal_precedence}, we can orient the edge so that \(X_{i+1}\) must be a parent of \(Y_j\) in \(\mathbb{G}\). Using Lemma \ref{lemma:mb_par}, we know that \(Par(Y_j) = \textbf{MB}(Y_j) \implies X_{i+1} \in \textbf{MB}(Y_j)\).
Thus \(X_{i+1} \in M^{i+1}_j\) which represent the \textbf{MB}(\(Y_j)\) for step \(i+1\).

Finally, \(\mathcal{M}^{i+1}_j\) still recovers the Markov Boundary of \(Y_j\) such that \[\forall U \in \{\boldsymbol{U'} - \mathcal{M}^{i+1}_j\}, Y_j \perp U|\mathcal{M}^{i+1}_j\]
\end{proof}

\section{Evaluation}
\subsection{Metrics}\label{appendix:eval}
The Precision, Recall, and F1-Score for Markov boundary estimation were computed as follows using the True set as the error pattern rule (True Markov Boundary) and the Inferred Markov Boundary set from OSCAR:

\begin{itemize}
    \item \textbf{Precision} (\(P\)) measures the proportion of correctly identified causal events among all inferred events:
    \[
    P = \frac{|\text{Inferred} \cap \text{True}|}{|\text{Inferred}|}
    \]
    where \(|\text{Inferred} \cap \text{True}|\) is the number of true positive causal events, and \(|\text{Inferred}|\) is the total number of inferred causal events.

    \item \textbf{Recall} (\(R\)) captures the proportion of correctly identified causal events among all true causal events:
    \[
    R = \frac{|\text{Inferred} \cap \text{True}|}{|\text{True}|}
    \]
    where \(|\text{True}|\) is the total number of true causal tokens.

    \item \textbf{F1-Score} (\(F_1\)) is the harmonic mean of precision and recall, providing a balanced measure:
    \[
    F_1 = \frac{2 \cdot P \cdot R}{P + R}
    \]
\end{itemize}
\subsection{PyCausalFS}
Local structure learning algorithms were all used with \(\alpha=0.1\) in the associated code: \url{https://github.com/wt-hu/pyCausalFS/tree/master/pyCausalFS/LSL}.
\subsection{MI-MCF}
MI-MCF \cite{mimcf} was used for comparison following the official implementation at \url{https://github.com/malinjlu/MI-MCF} we used \(\alpha = 0.05, L = 268, k_1 = 0.7, k_2=0.1\).
\section{Ablations}

\subsection{NADEs Quality.}\label{appendix:nades}
We did several ablations on the quality of the NADEs and their impact on the one-shot causal discovery phase. In particular, Table~\ref{tab:ablation_nades_comparison} presents multiple \(\text{Tf}_x, \text{Tf}_y\) with respectively 90 and 15 million parameters or 34 and 4 million parameters. We also varied the context window (conditioning set \(\boldsymbol{Z}\)), trained on different amounts of data (Tokens), and reported the classification results on the test set of \(\text{Tf}_y\) alone. We didn't output the Running time since it was always the same for all NADEs: \(1.27\) minutes of 50,000 samples and \(0.14\) for 5000.

\begin{table}[ht]
\centering
\caption{Ablations of the performance of Phase 1 (One-shot \textbf{MB} retrieval) in function of different NADEs with \(n=50{,}000\) and \(n=500\) samples averaged over 5-folds. Classification metrics use weighted averaging. Metrics are given in \(\%\).}
\label{tab:ablation_nades_comparison}
\begin{tabular}{lcccccc}
\toprule
\textbf{Tokens} & \textbf{Parameters} & \textbf{Context} & \textbf{Precision (↑)} & \textbf{Recall (↑)} & \textbf{F1 Score (↑)} & \textbf{Tfy F1 (↑)} \\
\midrule
\multicolumn{7}{c}{\textit{For \(n = 50{,}000\) samples}} \\
\midrule
1.5B & 105m & \(c = 4\)  & \(47.95 \pm 1.05\) & \(30.65 \pm 0.51\) & \(37.39 \pm 0.67\) & 88.6 \\
1.5B & 105m & \(c = 12\) & \(54.62 \pm 1.03\) & \(29.88 \pm 0.73\) & \(38.63 \pm 0.85\) & 90.43 \\
1.5B & 105m & \(c = 15\) & \(\mathbf{55.26 \pm 1.42}\) & \(\mathbf{31.37 \pm 0.82}\) & \(\mathbf{40.02 \pm 1.03}\) & 90.57 \\
1.5B & 105m & \(c = 20\) & \(49.52 \pm 1.59\) & \(\mathbf{31.76 \pm 0.85}\) & \(36.54 \pm 1.10\) & 91.19 \\
1.5B & 105m & \(c = 30\) & \(36.65\pm 1.18\) & \(22.75 \pm 0.78\) & \(26.57 \pm 0.91\) & \textbf{92.64} \\
300m & 47m & \(c = 20\) & \(39.49 \pm 1.77\) & \(26.30 \pm 0.89\) & \(29.01 \pm 1.10\) & 83.6 \\
\midrule
\multicolumn{7}{c}{\textit{For \(n = 500\) samples}} \\
\midrule
1.5B & 105m & \(c = 12\) & \(54.84 \pm 4.55\) & \(\mathbf{31.45 \pm 2.23}\) & \(\mathbf{39.95 \pm 2.83}\) & 90.43 \\
1.5B & 105m & \(c = 15\) & \(55.04 \pm 3.36\) & \(29.90 \pm 1.78\) & \(38.74 \pm 2.24\) & 90.57 \\
1.5B & 105m & \(c = 20\) & \(48.84 \pm 4.01\) & \(\mathbf{31.65 \pm 2.37}\) & \(36.19 \pm 2.65\) & \textbf{91.19} \\
300m & 47m & \(c = 20\) & \(38.23 \pm 2.91\) & \(25.31 \pm 2.39\) & \(27.92 \pm 2.25\) & 83.6 \\
\bottomrule
\end{tabular}
\end{table}
\subsection{Sampling Type}\label{abl:sampling}
We performed an ablation (Tab \ref{tab:sampling_comparison_transposed}) on the effect of sampling methods to estimate the expected value over all possible context \(\boldsymbol{Z}\). We used one A10 GPU on a sample of the test dataset (4000 random samples) composed of 205 labels with a batch size of 4 during inference.
We tested top-k sampling with \(k=\{20, 35\}\) \cite{fan-etal-2018-hierarchical} with and w/o a temperature scaler of \(T\) to log-probabilities \(\boldsymbol{\hat{x}}\) such that \[\boldsymbol{\hat{x}}' = \text{softmax}(\log{\boldsymbol{\hat{x}}}/T)\]

And a combination of top-k and a top-nucleus sampling \cite{Holtzman2020The} with different probability mass \(p=\{0.8, 1.2\}\) and finally a permutation of token position within the context c. 
We fixed a dynamic threshold with z score \(k=3\) and performed 10 runs. Then, we reported the average and standard deviation of each classification metric and elapsed time (sec).

Without a surprise, sampling increases the predictive performance of OSCAR by a large margin. More interestingly, different sampling types have different effects on specific averaging.
This has a 'smoothing' effect on the CMI curve when multiple labels are present in the sequence. When having no upsampling, the sensitivity of the CMI of different labels is increased, which makes it more difficult to capture a threshold and a potential cause. We can notice that globally, top-k sampling provides better results, especially with a combination of top-p=0.8 afterwards. 

Sampling with the same tokens (\textit{Permutation}) is not a good choice; sampling from the next-event prediction \(\text{Tf}_x\) yielded better results. We will choose \textbf{Top-k+p=0.8} for the increased F1 Micro and high F1 Macro, and Weighted.
\begin{table}[ht]
    \centering
    \caption{One-shot Classification performance and Elapsed Time (sec) across different sampling methods. Best results are shown in \textbf{Bold} and Best ex aequo in \uline{underline}.}
    \label{tab:sampling_comparison_transposed}
    \begin{tabular}{@{}lcccc@{}}
        \toprule
        Sampling Method & F1 Micro (\%) & F1 Macro (\%) & F1 Weighted (\%) & Time (sec) \\ \midrule
        w/o Sampling         & \(14.07\)                & \(12.29\)                & \(16.67\)                &  \(\mathbf{49.30 \pm 0.30}\)       \\
        Permutation & \( 18.22 \pm 0.36 \) & \( 13.75 \pm 0.09\) & \( 19.21\pm 0.03 \)         & \( 557.82 \pm 0.13 \) \\
        Top-k=20 & \( 26.77 \pm 0.71 \) & \( 23.83 \pm 0.19 \) & \( 29.25 \pm 0.07 \)         & \( 557.4 \pm 0.13  \) \\
        Top-k=35 & \(26.57 \pm 0.96 \) & \( 24.08 \pm 0.23 \) & \( 29.30 \pm 0.07 \)         & \( 557.35 \pm 0.10  \) \\
        Top-k=35+T=0.8 & \(27.36 \pm 0.65 \) & \( 23.77 \pm 0.21 \) & \( 28.98 \pm 0.07 \)         & \( 557.45 \pm 0.11  \) \\
         Top-k=35+T=1.2 & \(26.59 \pm 1.49 \) & \( \mathbf{24.62 \pm 0.29} \) & \( \mathbf{29.52 \pm 0.06} \) & \( 557.45 \pm 0.12  \) \\
          Top-k=25+p=0.8 & \( 27.98 \pm 0.67 \) & \( 23.82 \pm 0.28 \) & \( 29.18 \pm 0.07 \) & \( 558.07 \pm 0.07  \) \\
          \textbf{Top-k=35+p=0.8} & \(\mathbf{28.82 \pm 0.75}\) & \(24.06 \pm 0.25\) & \(29.17 \pm 0.07\)          & \(558.16 \pm 0.14\)  \\
          Top-k=35+p=0.9 & \(26.39 \pm 0.99\) & \(24.12 \pm 0.31\) & \(29.26 \pm 0.11\)          & \(558.11 \pm 0.12\)  \\
        Top-k=35+p=0.9+T=0.9 & \(27.63 \pm 0.75\) & \(23.90 \pm 0.24\) & \(29.04 \pm 0.09\)          & \(558.07 \pm 0.12\)  \\
        Top-k=35+p=0.9+T=1.1 & \(26.75 \pm 1.30\) & \(\mathbf{24.47 \pm 0.24}\) & \(29.45 \pm 0.09\)          & \(558.06 \pm 0.11\)  \\
        \bottomrule
    \end{tabular}
\end{table}

\subsection{Sampling Number}\label{sec:samplingnumber}
\begin{figure}[!h]
    \centering
    \caption{Evolution of several classification metrics (one-shot) and elapsed time per sample in function of the number of samples \(N\) chosen. Results are reported using a 1-sigma error bar.}
    \includegraphics[width=0.8\linewidth]{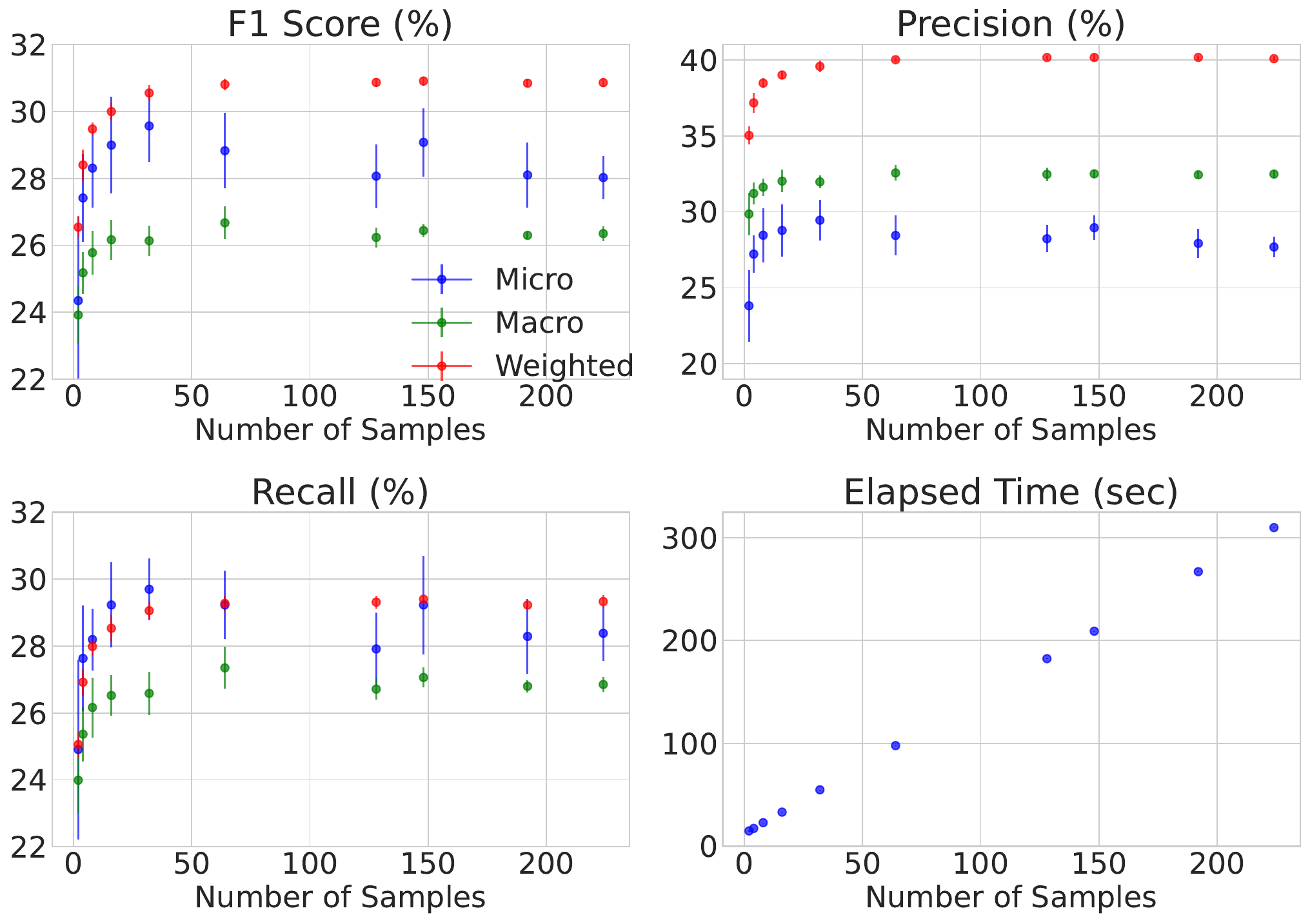}
    \label{abl:number_sampling}
\end{figure}

We experimented with different numbers of \(n\) for the sampling method across different averaging (micro, macro, weighted), Fig.~\ref{abl:number_sampling}. We performed 8 different runs and reported the average, standard deviation, and elapsed time. We can say that generally, sampling with a bigger \(N\) tends to decrease the standard deviation and give more reliable Markov Boundary estimation. Moreover, as we process more samples, the model is gradually improving at a logarithmic growth until it converges to a final score. We also verify that our time complexity is linear with the number of samples \(N\). 
Based on these results, we choose generally \(N=68 \) as the number of samples.

\subsection{Dynamic Thresholding}
\begin{figure}[!ht]
    \centering
    \caption{Evolution of one-shot F1 Score, Precision and Recall in function of coefficient \(k\). Results  are reported using 1-sigma error bar.}
    \label{abl:threshold_selection}
    \includegraphics[width=0.8\linewidth]{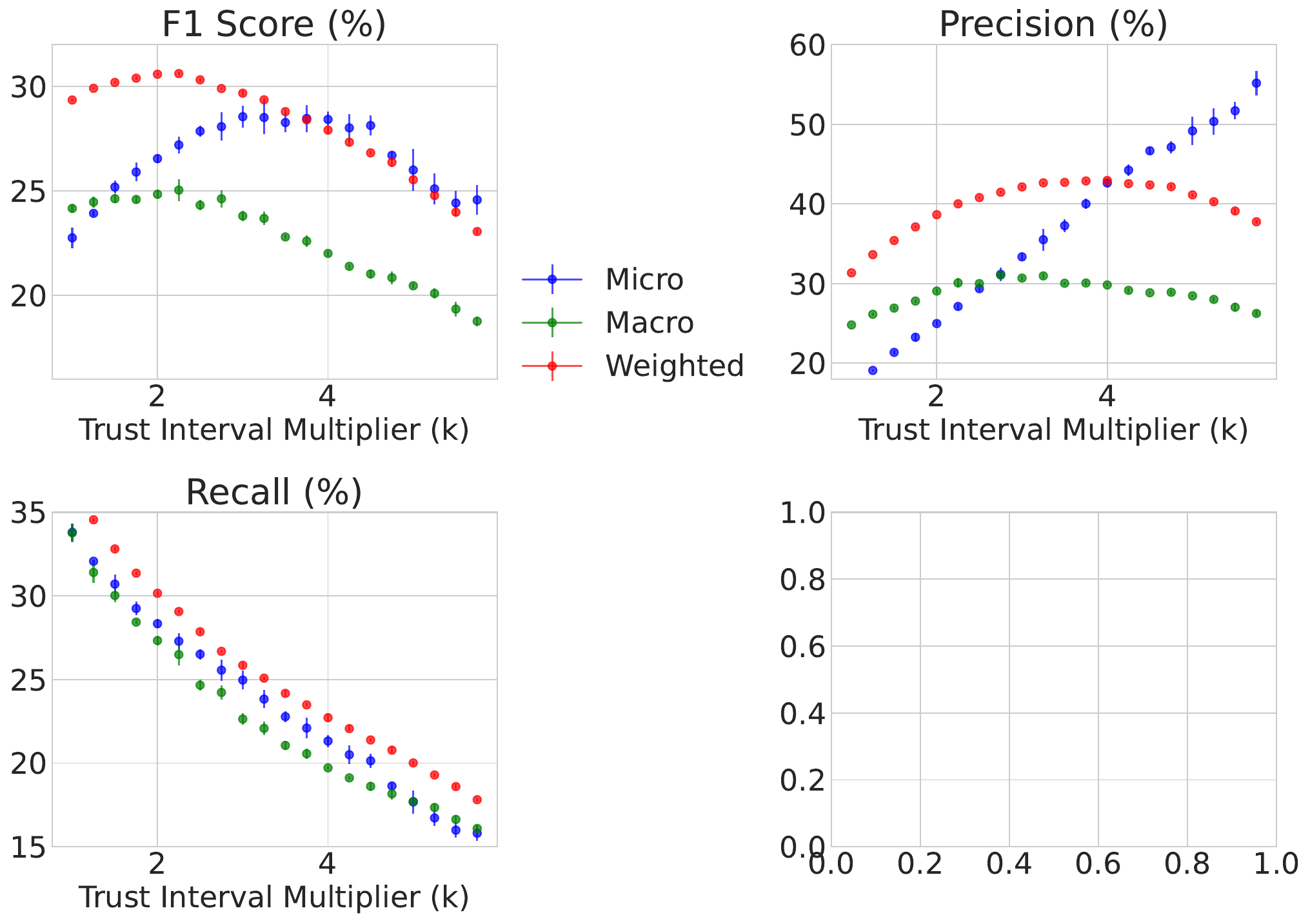}
\end{figure}

We performed ablations on the effect of \(k\) during the dynamic thresholding of the CMI (Eq.~\eqref{eq:cmi_epsilon}) to access conditional independence in Fig.~\ref{abl:threshold_selection}. To balance the classification metrics across the different averaging, we set \(k=2.75\).

\newpage
\section{Discussion on Assumptions}\label{sec:6_discussion}
Our approach relies on several assumptions that enable one-shot causal discovery under practical and computational constraints. 

\textbf{Temporal Precedence}
Temporal precedence (A\ref{assumption:temporal_precedence}) simplifies directionality and faithfulness to \(\mathbb{G}\). It allows for instantaneous influence, which aligns better with log-based data in cybersecurity or vehicle diagnostics, where events can co-occur at the same timestamp. However, this places strong reliance on precise event time-stamping. Even though we only test \(X_i \rightarrow Y_j\), this could falsify the conditioning test \(\boldsymbol{Z}\). 

\textbf{Bounded Lagged Effects.}
The bounded lagged effects (A\ref{assumption:lagged_effects}) assumption enables us to restrict causal influence and recover the \textbf{MB} of each label using Theorem~\ref{th:mb-recovery}. It also makes the computation faster. In most real-world sequences where relevant history is limited, this holds empirically. Nonetheless, in highly delayed causal chains, like financial transactions, some influence may be missed.

\textbf{Causal Sufficiency.}
As with many causal discovery approaches, we assume all relevant variables are observed (A\ref{assumption:causal_sufficiency}). Although it sounds like a strong assumption, interestingly, in high-cardinality domains such as vehicle diagnostics, the volume of recorded events may reduce but not eliminate the risk of hidden confounding.

\textbf{Inter-label Effects.}
By definition, the labels are explained solely by events. While simplifying multi-label causal discovery, this intrinsic assumption could be relaxed in future work by using the \textit{do} operator \cite{pearl_2009} to perform interventions on common causal variables of multiple labels. For example, our current framework estimates the Markov Boundaries for each label independently. However, inter-label dependencies can exist, particularly when labels share overlapping Markov Boundaries (e.g \(MB_1 =  [X_1, X_3], MB_2 = [X1, X_2]\). We propose to investigate a 'Phase 2' for OSCAR, focusing on inter-label dependencies through simulated interventions. For instance, if we consider a sequence \(S_1\) of two labels \(Y_1, Y_2\) with the MB above, we could perform counterfactual interventions by applying \(do(X_1=0), do(X_3=0)\)to \(S_1\). Then we would observe the average change in the likelihood of \(Y_1\)  which, if it is non-zero, would indicate a dependence between \(Y_1\) and \(Y_2\). \citet{learningcommoncausalvarlabel} points out that the assumptions of these inter-label dependencies are already anchored in the Markov Boundaries; we do the same here.

\textbf{NADEs.}
Due to the usage of flexible NADEs, we can relax common assumptions regarding data generation processes, such as Poisson Processes or SCMs.
Finally, as seen in the Ablations \ref{appendix:nades}, the effectiveness of OSCAR hinges on the capacity of \(\text{Tf}_x\) and \(\text{Tf}_y\) to approximate true conditional probabilities (A\ref{assumption:oracle}) and provide Oracle CI-test. While assuming Oracle tests are common in the literature \cite{mbb-by-mbb, feature_selection_review} and necessary to recover correct causal structures, this remains a strong assumption. And it is only valid to the extent that the models are perfectly trained. Especially for multi-label classification, performance may degrade in underrepresented regions of the data distribution, as we saw during the \textbf{MB} length comparison.

\section{Figures}

\begin{figure}[!h]
    \centering
\caption{An example of a causal graph extracted from a multi-label \textcolor{blue}{event} sequence where \textcolor{blue}{\(\text{MB}_1\)} represents the Markov Boundary of \textcolor{red}{\(Y_1\)} and \textcolor{cyan}{\(\text{MB}_2\)} the Markov Boundary of \textcolor{red}{\(Y_2\)}.}
\label{fig:markov_boundary_identification}
    \begin{tikzpicture}

        \draw[thick] (-2,0) -- (3.5,0) node[right] {Time};
        \draw[thick] (-2,0) -- (-2,2) node[right] {Event Type};

        \node[state,fill=blue] (x1) at (-1.5,0.5) {};
        \node[state,fill=blue] (x2) at (-0.5,0.5) {};
        \node[state,fill=blue] (x3) at (1,0.5) {};
        \node[state,fill=blue] (x4) at (2 ,0.5) {}; 

        \node[state,fill=red] (y) at (3,0.5) {}; 
        \node[state,fill=red] (y) at (3, 1) {}; 

        \draw[dashed] (-1.5,0) -- (-1.5,0.5);
        \draw[dashed] (-0.5,0) -- (-0.5,0.5);
        \draw[dashed] (1,0) -- (1,0.5);
        \draw[dashed] (2,0) -- (2,0.5);
        \draw[dashed] (3,0) -- (3,0.5);

        \node[below] at (-1.5,0) {\(X_1\)};
        \node[below] at (-0.5,0) {\(\cdots\)};

        \node[below] at (1,0) {\(X_{L-2}\)};
        \node[below] at (2,0) {\(X_{L-1}\)};

        \node[below] at (3,0) {\(\boldsymbol{Y}_{L}\)};

        \draw[thick,->] (4,-1) -- (5,-1) node[midway,above] {Identification};

        \node[state] (cx1) at (5,0) {\(X_{1}\)};
        \node[state] (cxd) at (7,0) {\(\cdots\)};
        \node[state] (cx2) at (9,0) {\(X_{L-2}\)};
        \node[state] (cxi) at (11,0) {\(X_{L-1}\)};
        \node[state] (cy) [above =of cxd] {\(Y_1\)};
        \node[state] (cy2) [above =of cxi] {\(Y_2\)};

        \path (cx1) edge (cxd);
        \path (cxd) edge  (cy);
        \path (cx1) edge[bend right=30] (cx2);
        \path (cx1) edge[bend right=30] (cxi);
        \path (cx2) edge (cxi);
        \path (cx2) edge[bend left=60] (cy2);
        \path (cxd) edge (cx2);
        \path (cxi) edge (cy2);
        \path (cxd) edge[bend left=60] (cxi);

        \node[draw=blue,dotted,fit=(cxd) (cxd), inner sep=0.2cm] (mb1) {};
        \node[anchor=south west, blue] at (mb1.north west) {\(\text{MB}_1\)};


        \node[draw=cyan,dotted,fit=(cx2) (cxi), inner sep=0.2cm] (mb2) {};
        \node[anchor=south west, cyan] at (mb2.north west) {\(\text{MB}_2\)};
    \end{tikzpicture}
\end{figure}
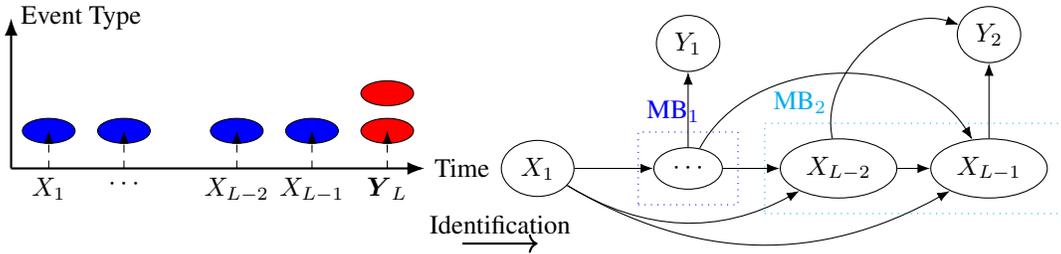

\section{Explanation example}\label{appendix:explaination_ex}
To enhance interpretability and illustrate the learned relationships, we present graphical explanations of error pattern occurrences based on sequences of Diagnostic Trouble Codes (DTCs). For each case, we selected representative samples that reflect diverse yet intuitive failure scenarios.

\begin{figure}[!h]
    \centering
    \caption{Example of a sequence of events (\textcolor{blue}{DTCs}) that lead to an emergency antenna dysfunction as outcome \textcolor{red}{labels}. The inhibitory strengths are shown in \textcolor{pink}{pink} and causal strengths in \textcolor{orange}{orange} and \textcolor{red}{red}}
    \includegraphics[width=0.75\linewidth]{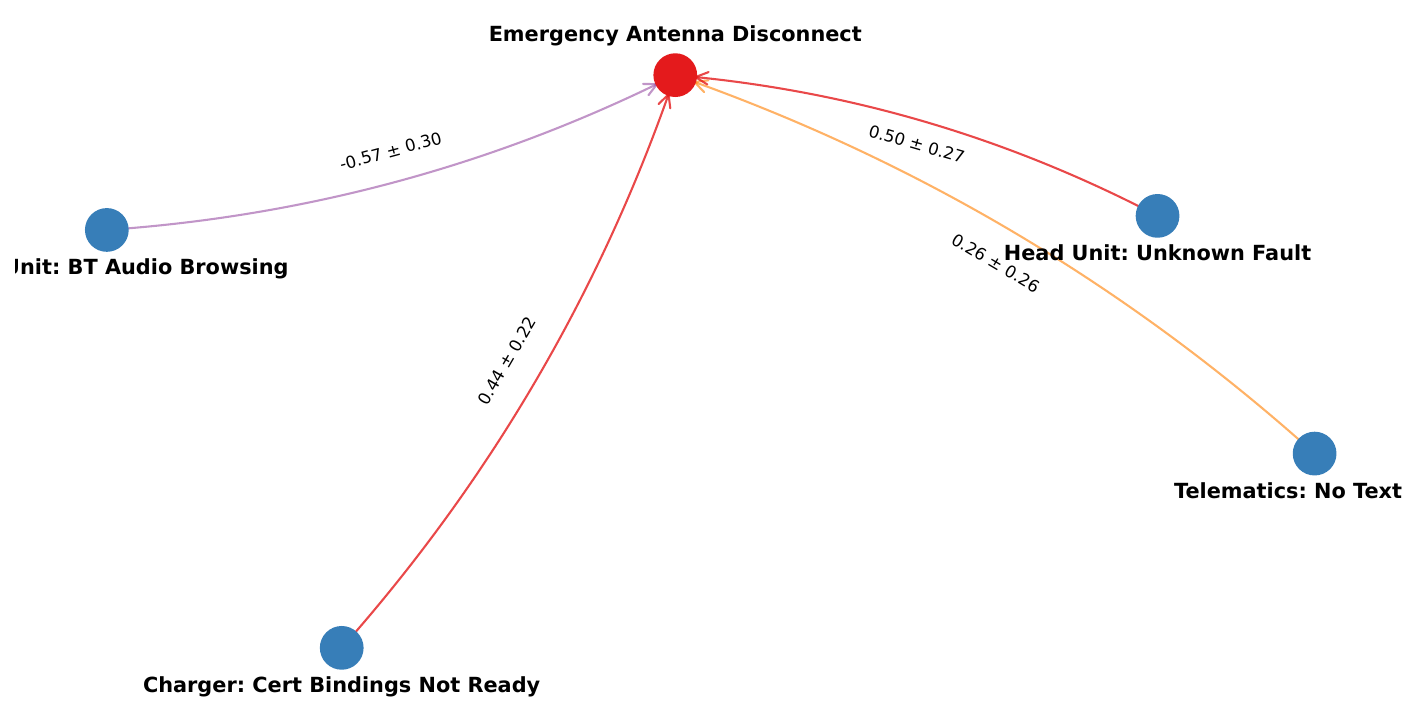}
    \label{fig:graph_cropped_antenna}
\end{figure}

\begin{figure}[!h]
    \centering
    \caption{Example of a sequence of events (\textcolor{blue}{DTCs}) that lead to an airbag and tire pressure malfunctions as outcome \textcolor{red}{labels}. The inhibitory strengths are shown in \textcolor{pink}{pink} and causal strengths in \textcolor{orange}{orange} and \textcolor{red}{red}}
    \includegraphics[width=1\linewidth]{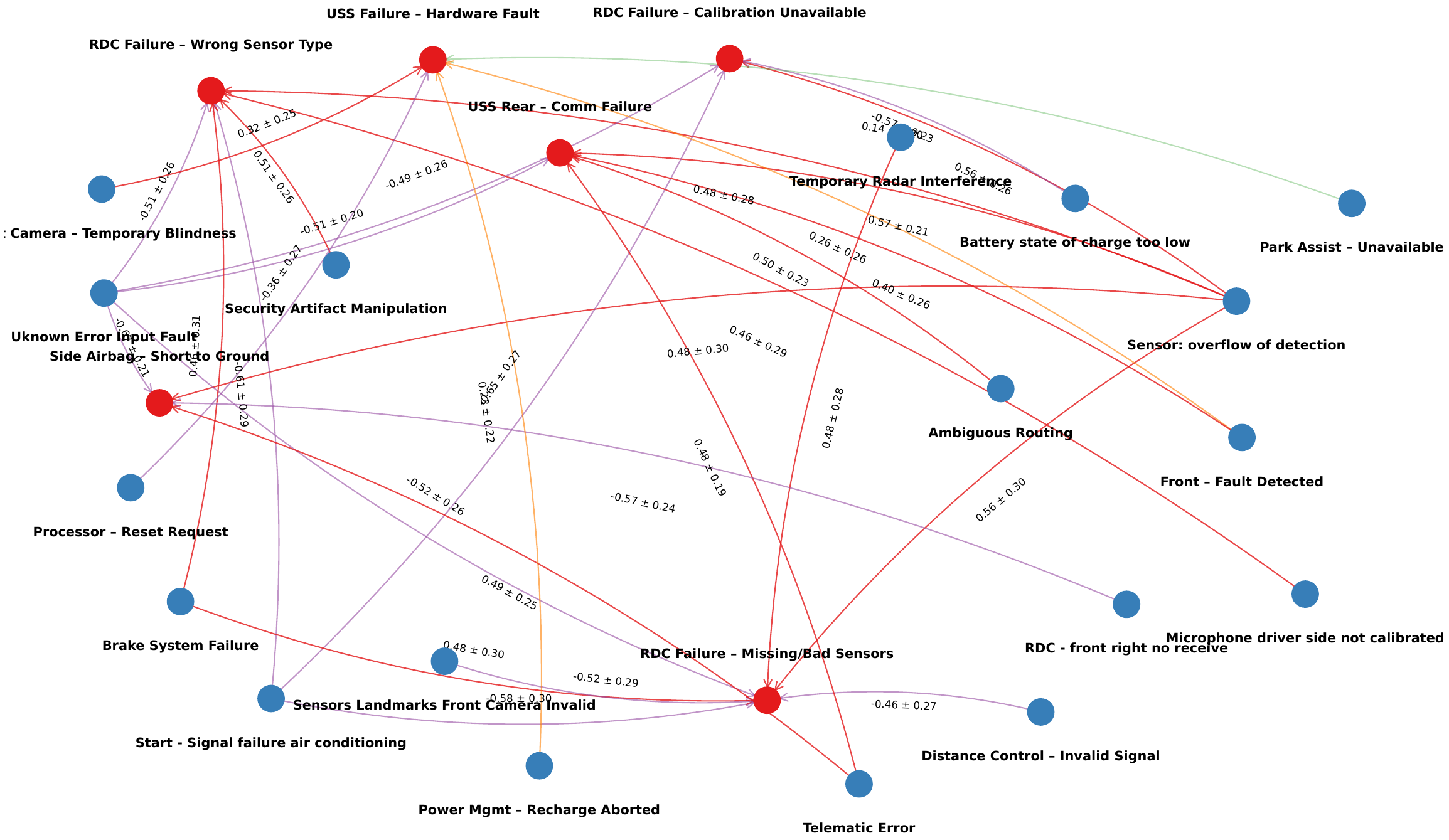}
    \label{fig:graph_cropped_airbag_rdc}
\end{figure}

Fig.~\ref{fig:graph_cropped_antenna} depicts a clear-cut example involving a single failure label related to the emergency antenna system. In contrast, Fig.~\ref{fig:graph_cropped_airbag_rdc} captures a more intricate case where airbag and tire pressure (RDC) malfunctions co-occur. These graphs highlight the influence of preceding events, with causal contributions shown in \textcolor{orange}{orange} and \textcolor{red}{red}, and inhibitory effects illustrated in \textcolor{pink}{pink}. Such visualisations serve to provide both human-understandable insights and support for the model’s reasoning process.

\begin{figure}[!h]
    \centering
    \caption{Example of a sequence of events (\textcolor{blue}{DTCs}) that lead to a steering wheel degradation and a power limitation as outcome \textcolor{red}{labels}. The inhibitory strengths are shown in \textcolor{violet}{violet} and causal strengths in \textcolor{orange}{orange} and \textcolor{red}{red} depending on the magnitude.}
    \includegraphics[width=0.8\linewidth]{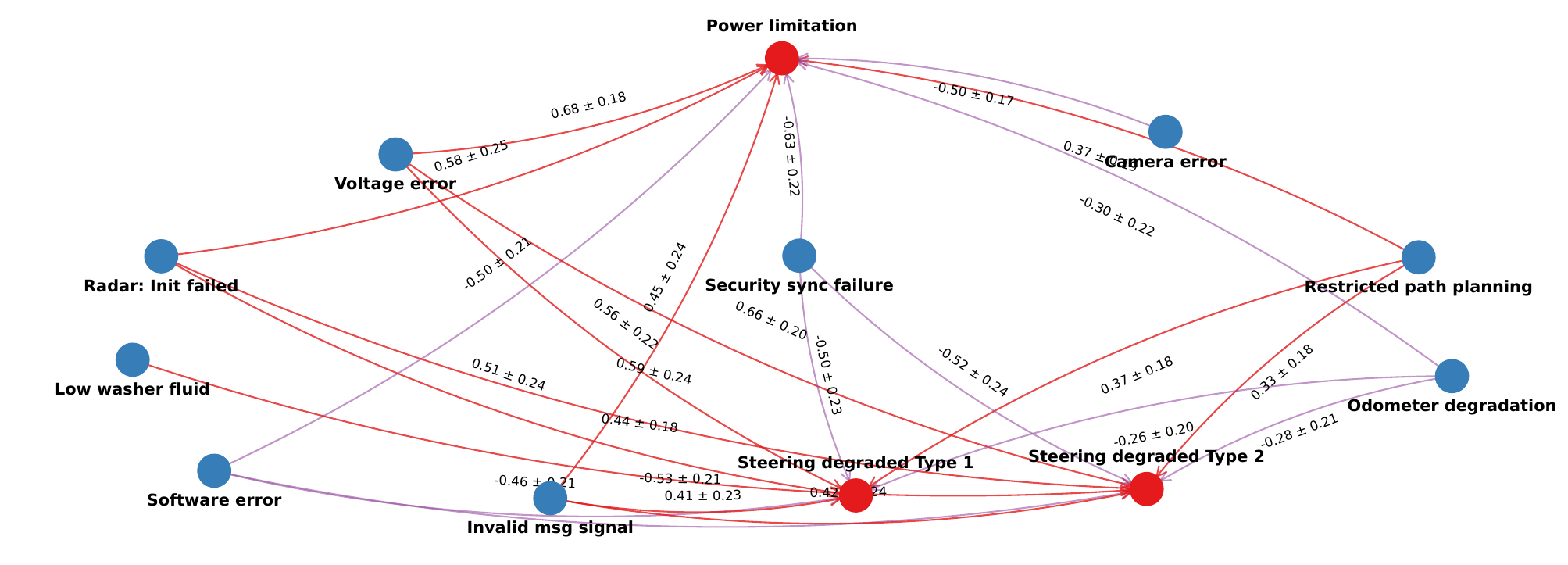}
    \label{fig:graph_cropped}
\end{figure}


\newpage
\newpage
\section{Implementation}
The following is the implementation of OSCAR in PyTorch \cite{pytorch}.
\begin{lstlisting}[language=Python, label={lst:oscar}]
def topk_p_sampling(z, prob_x, c: int, n: int = 64, p: float = 0.8, k: int = 35,
                       cls_token_id: int = 1, temp: float = None):
    # Sample just the context
    input_ = prob_x[:, :c]

    # Top-k first
    topk_values, topk_indices = torch.topk(input_, k=k, dim=-1)

    # Top-p over top-k values
    sorted_probs, sorted_idx = torch.sort(topk_values, descending=True, dim=-1)
    cum_probs = torch.cumsum(sorted_probs, dim=-1)
    mask = cum_probs > p
    
    # Ensure at least one token is kept
    mask[..., 0] = 0

    # Mask and normalize
    filtered_probs = sorted_probs.masked_fill(mask, 0.0)
    filtered_probs += 1e-8  # for numerical stability
    filtered_probs /= filtered_probs.sum(dim=-1, keepdim=True)

    # Unscramble to match the original top-k indices
    # Need to reorder the sorted indices back to the original top-k
    reorder_idx = torch.argsort(sorted_idx, dim=-1)
    filtered_probs = torch.gather(filtered_probs, -1, reorder_idx)

    batched_probs = filtered_probs.unsqueeze(1).repeat(1, n, 1, 1)        # (bs, n, seq_len, k)
    batched_indices = topk_indices.unsqueeze(1).repeat(1, n, 1, 1)        # (bs, n, seq_len, k)

    sampled_idx = torch.multinomial(batched_probs.view(-1, k), 1)         # (bs*n*seq_len, 1)
    sampled_idx = sampled_idx.view(-1, n, c).unsqueeze(-1)

    sampled_tokens = torch.gather(batched_indices, -1, sampled_idx).squeeze(-1)
    sampled_tokens[..., 0] = cls_token_id

    # Reconstruct full sequence
    z_expanded = z.unsqueeze(1).repeat(1, n, 1)[..., c:]
    return torch.cat((sampled_tokens, z_expanded), dim=-1)

from torch import nn
def OSCAR(tfe: nn.Module, tfy: nn.Module, batch: dict[str, torch.Tensor], c: int, n: int, eps: float=1e-6, topk: int=20, k: int=2.75, p=0.8) -> torch.Tensor:
    """ tfe, tfy: are the two autoregressive transformers (event type and label)
        batch: dictionary containing a batch of input_ids and attention_mask of shape (bs, L) to explain.
        c: scalar number defining the minimum context to start inferring, also the sampling interval.
        n: scalar number representing the number of samples for the sampling method.
        eps: float for numerical stability
        topk: The number of top-k most probable tokens to keep for sampling
        k: Number of standard deviations to add to the mean for dynamic threshold calculation
        p: Probability mass for top-p nucleus
    """
    o = tfe(attention_mask=batch['attention_mask'], input_ids=batch['input_ids'])['prediction_logits'] # Infer the next event type
    x_hat = torch.nn.functional.softmax(o, dim=-1)

    b_sampled = topk_p_sampling(batch['input_ids'], x_hat, c, k=topk, n=n, p=p) # Sampling up to (bs, n, L)
    n_att_mask = batch['attention_mask'].unsqueeze(1).repeat(1, n, 1)

    with torch.inference_mode():
        o = tfy(attention_mask=n_att_mask.reshape(-1, b_sampled.size(-1)), input_ids=b_sampled.reshape(-1, b_sampled.size(-1))) # flatten and infer
        prob_y_sampled = o['ep_prediction'].reshape(b_sampled.size(0), n, batch['input_ids'].size(-1)-c, -1) # reshape to (bs, n, L-c)

        # Ensure probs are within (eps, 1-eps)
        prob_y_sampled = torch.clamp(prob_y_sampled, eps, 1 - eps)

        y_hat_i = prob_y_sampled[..., :-1, :] # P(Yj|z)
        y_hat_iplus1 = prob_y_sampled[..., 1:, :] # P(Yj|z, x_i) 

        # Compute the CMI & CS and average across sampling dim
        cmi = torch.mean(y_hat_iplus1*torch.log(y_hat_iplus1/y_hat_i)+ (1-y_hat_iplus1)*torch.log((1-y_hat_iplus1)/(1-y_hat_i)), dim=1)
        # (BS, L, Y)
        cs = y_hat_iplus1 - y_hat_i
        cs_mean = torch.mean(cs, dim=1)
        cs_std = torch.std(cs, dim=1)

        # Confidence interval for threshold
        mu = cmi.mean(dim=1)
        std = cmi.std(dim=1)
        dynamic_thresholds = mu + std * k

        # Broadcast to select an individual dynamic threshold
        cmi_mask = cmi >= dynamic_thresholds.unsqueeze(1)

        cause_token_indices = cmi_mask.nonzero(as_tuple=False)
        # (num_causes, 3) --> each row is [batch_idx, position_idx, label_idx]
        return cause_token_indices, cs_mean, cs_std, cmi_mask
\end{lstlisting}

\begin{remark}
    Since \textit{tfy} contains tfe as backbone, in practice we need only one forward pass from tfy and extract also \(\hat{\boldsymbol{x}}\), so \textit{tfe} is not needed. We let it to improve understanding and clarity.
\end{remark}

\end{document}